\documentclass[
]{ceurart}

\usepackage{listings}
\begin{document}

\copyrightyear{2025}
\copyrightclause{Copyright for this paper by its authors.
  Use permitted under Creative Commons License Attribution 4.0
  International (CC BY 4.0).}


\conference{IberLEF 2025, September 2025, Zaragoza, Spain}

\title{Classification of Hope in Textual Data using Transformer-Based Models}


\author[1]{Chukwuebuka Fortunate Ijezue}[%
email=cijezue@ttu.edu,
url=https://ijezue.github.io/site/,
]

\author[1]{Fredrick Eneye Tania-Amanda Nkoyo}[%
email=tafredri@ttu.edu,
url=https://crystal4000.github.io/academic\_portfolio/,
]
\cormark[2]

\author[1]{Maaz Amjad}[%
email=maaz.amjad@ttu.edu,
url=https://maazamjad.com/,
]

\fnmark[1]
\address[1]{Department of Computer Science, Texas Tech University, Lubbock, Texas, United States}

\cortext[2]{Corresponding author.}
\fntext[1]{These authors contributed equally.}

\begin{abstract}
This paper presents a transformer-based approach for classifying hope expressions in text. We developed and compared three architectures (BERT, GPT-2, and DeBERTa) for both binary classification (``Hope" vs. ``Not Hope") and multiclass categorization (five hope-related categories). Our initial BERT implementation achieved 83.65\% binary and 74.87\% multiclass accuracy. In the extended comparison, BERT demonstrated superior performance (84.49\% binary, 72.03\% multiclass accuracy) while requiring significantly fewer computational resources (443s vs. 704s training time) than newer architectures. GPT-2 showed lowest overall accuracy (79.34\% binary, 71.29\% multiclass), while DeBERTa achieved moderate results (80.70\% binary, 71.56\% multiclass) but at substantially higher computational cost (947s for multiclass training). Error analysis revealed architecture-specific strengths in detecting nuanced hope expressions, with GPT-2 excelling at sarcasm detection (92.46\% recall). This study provides a framework for computational analysis of hope, with applications in mental health and social media analysis, while demonstrating that architectural suitability may outweigh model size for specialized emotion detection tasks.
\end{abstract}

\maketitle

\section{Introduction}
In natural language processing (NLP), the computational analysis of text's emotive and emotional content is becoming a prominent area of study. Sentiment analysis \cite{hu2004mining}, emotion detection\cite{mohammad2016sentiment}, and toxicity classification \cite{ZhangDetecting} have all seen substantial research, but the particular field of hope detection and classification remains relatively unexplored. As a complex psychological concept, hope is essential to social discourse \cite{herth1992hope}, mental health \cite{snyder2002hope}, and human communication \cite{snyder2002hope}. Automatically identifying and classifying hopeful textual statements has potential applications in crisis response \cite{li2019survey}, social media analysis, political discourse analysis \cite{liu2022sentiment}, and mental health monitoring \cite{coppersmith2014quantifying}.

This study presents a comprehensive approach to hope classification using transformer-based deep learning models. We developed a two-tiered classification system: (1) a binary classifier that distinguishes between hopeful expressions and those that are not, and (2) a multiclass classifier that categorizes text into five distinct hope-related categories: Not Hope, Generalized Hope, Realistic Hope, Unrealistic Hope, and Sarcasm. By differentiating between various forms of hopeful expressions, this granular approach enables a more thorough understanding of how hope manifests in text.

Our research begins with implementing BERT (Bidirectional Encoder Representations from Transformers) \cite{devlin2019bert} for hope classification, leveraging its contextual understanding capabilities that have shown state-of-the-art performance on various NLP tasks. We then expand our investigation to compare BERT with more advanced transformer architectures: GPT-2, which employs unidirectional attention and benefits from a larger pretraining corpus, and DeBERTa, which utilizes a disentangled attention mechanism designed to better capture semantic nuances. 

This comprehensive comparison addresses a critical question in affective computing: do newer, more complex language models provide meaningful performance improvements for specialized emotional detection tasks like hope classification? Our experimental results reveal interesting patterns, with BERT achieving the highest accuracy in both binary (84.49\%) and multiclass (72.03\%) tasks, despite being architecturally simpler than alternatives. While DeBERTa (80.70\% binary, 71.56\% multiclass) and GPT-2 (79.34\% binary, 71.29\% multiclass) showed competitive performance, they required substantially higher computational resources, with DeBERTa taking more than twice the training time of BERT for multiclass classification. Notably, GPT-2 demonstrated particular strength in detecting sarcastic expressions of hope.

These findings suggest that model complexity does not necessarily correlate with performance improvement for hope classification tasks, highlighting the importance of architecture-task alignment in emotion detection systems. By evaluating model accuracy, training efficiency, and error patterns across these architectures, we identify the optimal approach for hope detection in practical applications, balancing performance against computational requirements. This research contributes to the growing field of affective computing by providing empirical evidence on the relative efficacy of different transformer architectures for the specialized task of hope classification, while establishing a framework for computational analysis of hope with applications in mental health and social media analysis.

\section{Literature Review}

\subsection{Hope in Computational Linguistics}
Hope speech detection is an emerging area within Natural Language Processing (NLP) that focuses on identifying and distinguishing encouraging, supportive, and positive content, contrasting with the more established domain of hate or offensive speech detection. While sentiment analysis has been extensively studied \cite{hu2004mining}, nuanced emotions such as hope remain comparatively underexplored. \citet{chakravarthi2022overview} define hope speech as messages that “offer support, reassurance, suggestions, inspiration and insight” to foster optimism. Foundational psychological frameworks by \citet{snyder2002hope} continue to inform computational approaches to modeling hope.

A significant milestone in this area was the release of the HopeEDI dataset by \citet{chakravarthi2020hopeedi}, comprising 28,451 English, 20,198 Tamil, and 10,705 Malayalam YouTube comments annotated for hope speech. This multilingual corpus served as the basis for the first shared task on hope speech detection at the Workshop on Language Technology for Equality, Diversity, and Inclusion (LT-EDI) in 2021. In this task, \citet{saumya-mishra-2021-iiit} established early classification baselines using classical machine learning and neural models.

Subsequent studies have expanded both the scope and sophistication of computational hope classification. \citet{DBLP:journals/jksucis/MalikNMI23} introduced a new English–Russian dataset and explored cross-lingual training. Their RoBERTa-based model demonstrated that translating English content into Russian before training could achieve 94\% accuracy and an F1 score of 80.2\% on binary hope classification—highlighting the efficacy of transfer learning in low resource languages.

Although earlier approaches predominantly relied on classical machine learning algorithms such as Support Vector Machines (SVM), Logistic Regression, and K-Nearest Neighbors (KNN) with TF–IDF features \cite{Yigezu2023Multilingual}, recent research have shifted toward deep learning. \citet{saumya-mishra-2021-iiit} employed convolutional and recurrent networks, but by 2022–2023, transformer-based models emerged as state-of-the-art. For instance, baseline experiments using XLM-RoBERTa on the English HopeEDI dataset demonstrated its superiority over traditional classifiers. RoBERTa achieved a weighted F1 score of approximately 0.93, compared to ~0.90 for KNN and ~0.87 for SVM \cite{Chakravarthi2022Multilingual}. In terms of macro-F1, it reached 0.52, outperforming KNN (0.40) and SVM (0.32). Similarly, \citet{DBLP:journals/jksucis/MalikNMI23} found that a fine-tuned RuBERTa model using a translation-based training pipeline consistently outperformed baseline methods.

\subsection{Transformer Architectures for Emotion Detection}
BERT \cite{devlin2019bert} introduced bidirectional context modeling and has achieved state-of-the-art results across various NLP tasks. Its bidirectional attention mechanism allows it to consider the full context when classifying emotional content. GPT-2 \cite{radford2019language} employs unidirectional attention but benefits from a larger pre-training corpus, potentially capturing more linguistic patterns related to hope expressions. DeBERTa \cite{hedeberta} enhances BERT with disentangled attention, separately computing content and position information, which theoretically improves contextual understanding of complex emotions.

\subsection{Comparative Performance Studies}
Comparative analyses of transformer architectures have shown task-dependent performance variations. While newer models often outperform older ones on general benchmarks, specialized tasks may reveal different patterns. \citet{gao-etal-2023-small} demonstrated that small pre-trained language models can be fine-tuned to match larger models' performance, suggesting that model architecture and fine-tuning strategies are crucial for task-specific performance.

\section{Methodology}
This section details our comprehensive approach to hope classification, covering our dataset characteristics, pre-processing strategies, model architectures, implementation details, and evaluation framework. We present both our original BERT implementation and the extended comparison of three transformer architectures (BERT, GPT-2, and DeBERTa) to provide a thorough analysis of hope detection capabilities.
\subsection{Dataset}
This study employs custom datasets for hope classification, obtained from the PolyHope shared task \cite{balouchzahi2023polyhope, sidorov2023regret, garcia2023lgtb, garcia2024iberlef, jimenez2023hope, chakravarthi2020hopeedi, chakravarthi2022overview, butt2025overview, butt2025optimism, sidorov2024mind, balouchzahi2025urduhope} at IberLEF 2025 \cite{iberlef2025overview}. The training dataset contained 5,233 samples, while the development/test dataset comprised 1,902 samples. Both datasets maintained similar class distributions, ensuring consistency between training and evaluation. The dataset supports two classification schemes: a binary task (``Hope'' vs. ``Not Hope'') and a multiclass task with five categories (``Not Hope,'' ``Generalized Hope,'' ``Realistic Hope,'' ``Unrealistic Hope,'' and ``Sarcasm''). For the binary classification, the training set contained 2,426 (46.36\%) ``Hope'' samples and 2,807 (53.64\%) ``Not Hope'' samples, with the test set maintaining a similar distribution of 899 (47.27\%) ``Hope'' and 1,003 (52.73\%) ``Not Hope'' samples. The multiclass distribution was also consistent across both sets, with the following breakdown in the training data: ``Not Hope'' (42.90\%), ``Generalized Hope'' (24.54\%), ``Sarcasm'' (13.22\%), ``Realistic Hope'' (10.32\%), and ``Unrealistic Hope'' (9.02\%). The test set maintained nearly identical proportions. This balanced representation across classes helped ensure the models could learn to distinguish between all categories effectively. The text samples varied in length from 24 to 886 characters, with an average length of approximately 188 characters. This variation in text length provided the models with diverse linguistic patterns and expressions of hope across different contexts. To ensure robust model development, we implemented an 80-20 train-validation split on the training data, maintaining the same random seed (42) across experiments for consistency and reproducibility.

\subsection{Text Pre-processing}
In this study, our approach to text pre-processing differed between the original implementation and the extended comparison. In the original BERT implementation, we deliberately minimized pre-processing, feeding raw text directly into the tokenization pipeline to leverage BERT's capability to capture contextual nuances. For the extended comparison between models, we applied basic text cleaning through a custom function that converted text to lowercase, removed URLs and web links, removed hashtags and user mentions (patterns like \#word and @word), and removed punctuation. This cleaned text was stored in a separate 'clean\_text' column and used for tokenization across all three models. This standardized preprocessing in the extended comparison ensured a fair evaluation across different transformer architectures while allowing us to assess whether specialized cleaning benefits these pre-trained models. The contrast between approaches also enabled us to evaluate the impact of pre-processing on model performance for hope classification tasks.

\subsection{Transformer Model Architectures}
Our study implements and compares three state-of-the-art transformer architectures for hope classification, with BERT used in our original implementation and all three models (BERT, GPT-2, and DeBERTa) compared in our extended analysis.

\subsubsection{BERT Architecture}
In both our original and extended implementations, we utilized the `bert-base-uncased' variant from Hugging Face's Transformers library. This BERT model comprises 12 transformer layers, 12 attention heads, and 768 hidden dimensions, totaling approximately 110 million parameters. BERT's bidirectional attention mechanism enables the model to consider the full context when representing each word, potentially beneficial for capturing complex hope expressions. In our original implementation, BERT served as the sole architecture for establishing baseline performance in hope classification.

\subsubsection{GPT-2 Architecture}
For our extended comparison, we incorporated the GPT-2 base model (124M parameters) with its autoregressive architecture. Unlike BERT's bidirectional attention, GPT-2 uses unidirectional attention where each token can only attend to previous tokens in the sequence. While this limitation might affect classification performance, GPT-2's larger pre-training corpus potentially provides richer semantic representations beneficial for hope classification. Special consideration was required for GPT-2 implementation, including setting the pad token to match the EOS token and disabling the cache to avoid errors during training.

\subsubsection{DeBERTa Architecture}
Also included only in our extended comparison, DeBERTa (base version, 140M parameters) implements a disentangled attention mechanism that separately computes attention weights for content and position information. This approach theoretically allows for more nuanced contextual understanding, potentially beneficial for distinguishing between subtle variations of hope expressions and identifying sarcasm. DeBERTa represents the most complex architecture in our comparison, offering insights into whether architectural sophistication translates to improved hope classification performance.

\subsection{Model Implementation and Training}
Our technical approach evolved across the original and extended studies, with consistent use of the Hugging Face Transformers library and TensorFlow backend throughout both phases.

\subsubsection{Original BERT Implementation}
In our initial implementation, we focused exclusively on BERT using \textbf{TFBertForSequenceClassification} for both binary and multiclass classification tasks. We employed the BERT tokenizer with a maximum sequence length of 128 tokens, with padding and truncation applied as needed. The model was compiled with Adam optimizer (learning rate 2e-5) and SparseCategoricalCrossentropy loss function. We trained the model for 3 epochs with a batch size of 8, using accuracy as our primary evaluation metric. Model checkpointing was implemented to save the best-performing model based on validation accuracy.

\subsubsection{Extended Implementation Comparison}
For our comparative analysis, we expanded to include all three transformer architectures, implementing custom setup functions for each model. For tokenization, each model used its corresponding tokenizer with consistent parameters: maximum sequence length of 128 tokens, padding enabled, and truncation applied. For GPT-2, which lacks a dedicated pad token, we assigned the EOS token as the pad token and set \texttt{use\_cache=False} to prevent errors with the \texttt{past\_key\_values} parameter. Additionally, GPT-2 required inputs structured as dictionaries, necessitating the use of TensorFlow Dataset API for compatibility.

Across all models in our extended comparison, we maintained the same optimizer (Adam), loss function (SparseCategoricalCrossentropy), and learning rate (2e-5) while increasing the batch size to 16. Each model was trained for 3 epochs with identical ModelCheckpoint callbacks to ensure fair comparison of architectural differences rather than training hyperparameters. This standardized approach helped isolate architectural performance differences while mitigating overfitting through validation-based checkpointing. All models were saved in TensorFlow format for consistency and to facilitate deployment and further experimentation.

\subsection{Evaluation Framework}
Our evaluation strategy remained consistent across both the original BERT implementation and extended model comparison. We primarily relied on accuracy as our main metric for overall performance assessment, allowing direct comparison between models and with prior research in hope classification. Additionally, we calculated precision, recall, and F1 scores for both weighted and macro averages to provide a more nuanced understanding of model performance across classes.
For the extended comparison, we expanded our analysis to include training time as a measure of computational efficiency, an important consideration for real-world deployment scenarios. We also generated confusion matrices for each model, revealing specific classification patterns and highlighting each architecture's strengths and weaknesses in distinguishing between different hope categories, particularly their ability to identify subtle distinctions between hope types and sarcasm.

\subsection{Computational Environment}
Our study employed different computational resources across implementation phases. For the original BERT implementation, we utilized Texas Tech University's High-Performance Computing Center (HPCC) with NVIDIA A100 GPUs (40GB memory), providing substantial computational power for baseline model development. For the extended comparison, we shifted to Google Colab with NVIDIA T4 GPUs, which offered more accessibility while still providing sufficient capacity for comparative analysis. This environment change explains some of the training time differences observed between implementations. Both environments used Python 3.x with TensorFlow as the primary framework, supplemented by Hugging Face Transformers library, pandas for data manipulation, and scikit-learn for evaluation metrics. Despite the different GPU types, we maintained consistent training parameters and evaluation protocols to ensure meaningful comparisons across models and implementations.

\section{Results}
\subsection{Comparison of Original and Extended Implementations}
Table~\ref{tab:performance_comparison} shows the performance metrics of our original BERT implementation and the extended model comparison study. In our original implementation, BERT achieved 83.65\% accuracy for binary classification and 74.87\% for multiclass classification. The extended implementation yielded different results across models, with refined BERT showing notable improvement in binary classification (84.49\% vs. 83.65\%) but a decrease in multiclass performance (72.03\% vs. 74.87\%).

This performance difference between implementations can be attributed to several factors. First, the preprocessing approach differed, with the extended study applying more comprehensive text cleaning. We observed a drop in multiclass classification accuracy for BERT, despite other architectural and training conditions being the same. This accuracy decline suggests that the additional cleaning may have removed important linguistic features such as capitalization, punctuation-based emphasis, or hashtags that contribute to the nuanced expression of hope. This is in line with the findings of \citet{siino2024text_preprocessing}, who showed that in some cases minimal preprocessing like lowercasing can reduce the performance of transformer-based models.

Second, the computational environments varied (HPCC A100 GPUs vs. Google Colab T4 GPUs), potentially affecting optimization during training. Finally, the batch size increased from 8 in the original implementation to 16 in the extended comparison, which may have affected the learning dynamics.
It's particularly interesting that the multiclass performance declined across all models in the extended implementation (ranging from 71.29\% to 72.03\%) compared to our original BERT implementation (74.87\%). This consistent decrease suggests that either the original implementation benefited from a particularly advantageous random initialization or data split, or that the text pre-processing applied in the extended study may have removed linguistic features valuable for distinguishing between nuanced hope categories.

\begin{table}[ht]
    \centering
    \caption{Performance Comparison Between Original and Extended Model Implementations}
    \label{tab:performance_comparison}
    \begin{tabular}{|l|c|c|c|c|c|c|c|c|}
        \hline
        \textbf{Model} & \textbf{W-Prec} & \textbf{W-Rec} & \textbf{W-F1} & \textbf{M-Prec} & \textbf{M-Rec} & \textbf{M-F1} & \textbf{Acc} & \textbf{Time (s)} \\
        \hline
        \multicolumn{9}{|c|}{\textit{Binary Classification}} \\
        \hline
        Original BERT & 0.842 & 0.837 & 0.837 & 0.839 & 0.839 & 0.837 & 0.837 & --- \\
        Extended BERT & 0.845 & 0.845 & 0.845 & 0.844 & 0.845 & 0.845 & 0.845 & 443 \\
        Extended GPT-2 & 0.824 & 0.793 & 0.791 & 0.818 & 0.801 & 0.792 & 0.793 & 527 \\
        Extended DeBERTa & 0.829 & 0.807 & 0.805 & 0.824 & 0.813 & 0.806 & 0.807 & 704 \\
        \hline
        \multicolumn{9}{|c|}{\textit{Multiclass Classification}} \\
        \hline
        Original BERT & 0.776 & 0.749 & 0.752 & 0.714 & 0.753 & 0.719 & 0.749 & --- \\
        Extended BERT & 0.761 & 0.720 & 0.731 & 0.691 & 0.723 & 0.693 & 0.720 & 539 \\
        Extended GPT-2 & 0.718 & 0.713 & 0.711 & 0.669 & 0.667 & 0.662 & 0.713 & 530 \\
        Extended DeBERTa & 0.724 & 0.716 & 0.713 & 0.694 & 0.685 & 0.679 & 0.716 & 948 \\
        \hline
    \end{tabular}
\end{table}

\begin{figure}[ht]
  \centering
  \begin{minipage}{\textwidth}
    \centering
    \includegraphics[width=0.75\textwidth]{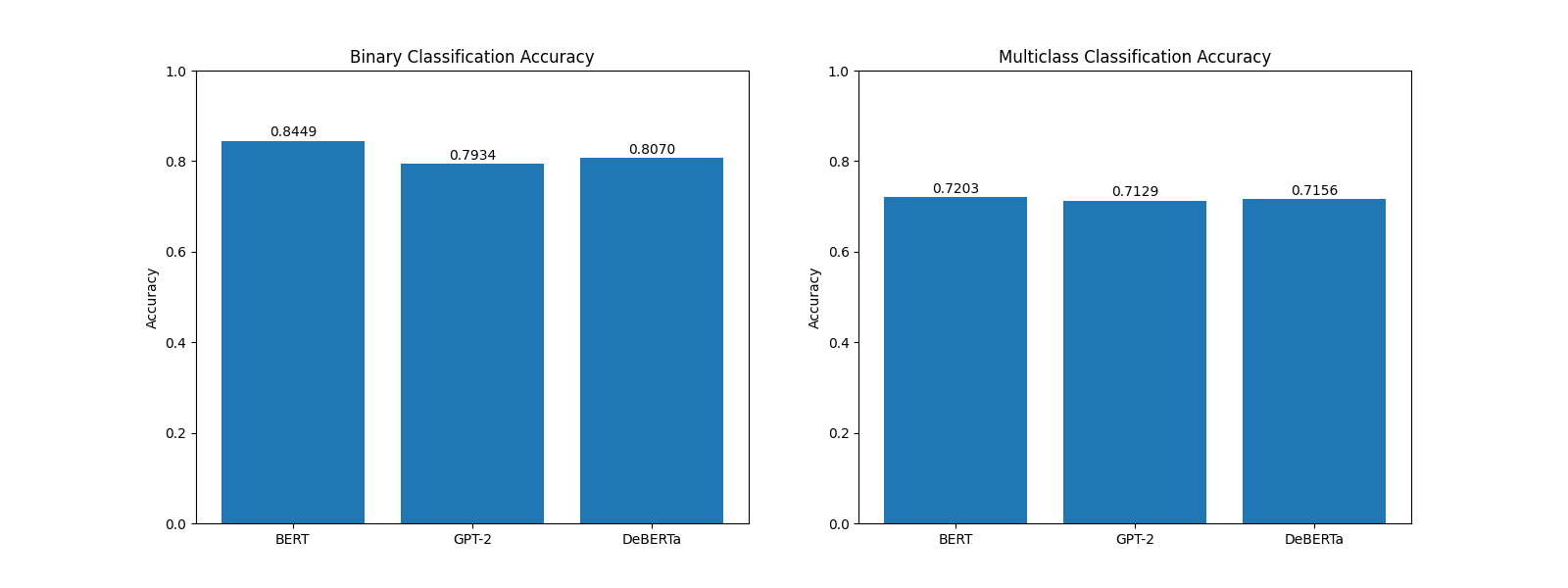}
    \caption{Accuracy comparison across models for binary and multiclass hope classification tasks.}
    \label{fig:accuracy}
  \end{minipage}
  
  \vspace{1cm} 
  
  \begin{minipage}{\textwidth}
    \centering
    \includegraphics[width=0.75\textwidth]{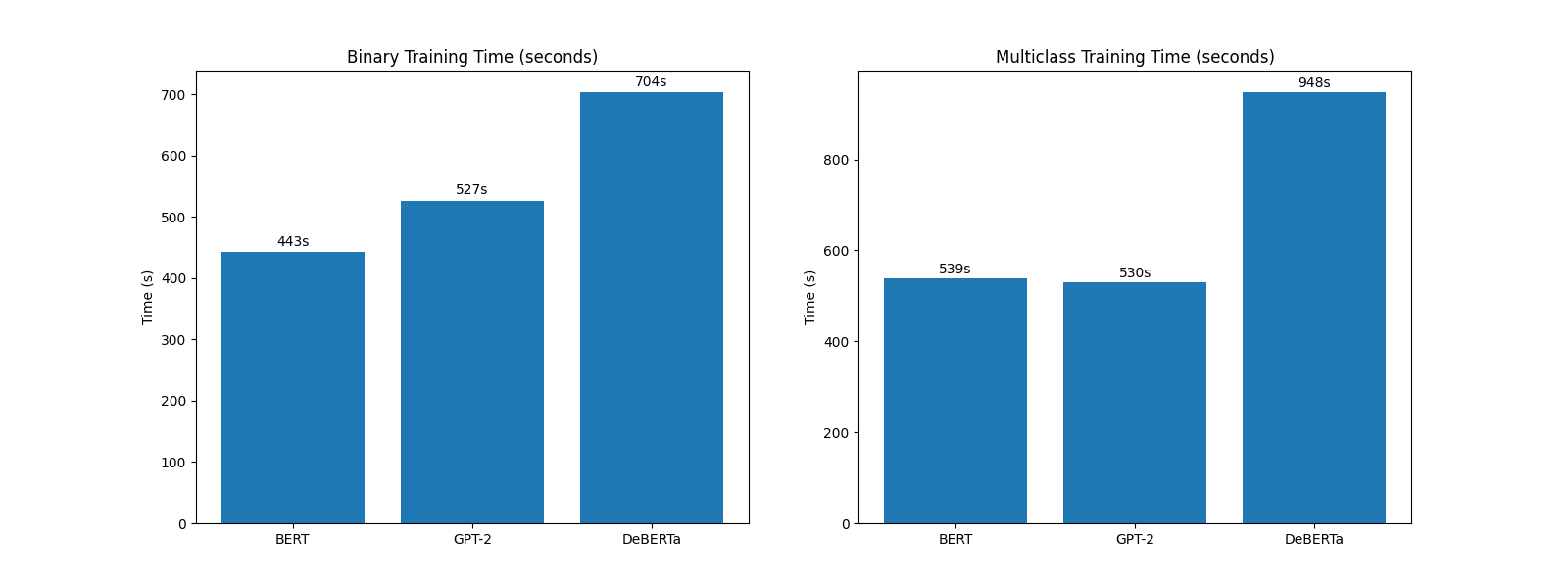}
    \caption{Training time comparison across models.}
    \label{fig:training_time}
  \end{minipage}
\end{figure}

\subsection{Model Performance Comparison}
Building upon our comparison with the original BERT implementation, we examined the relative performance of our three models in the extended study as shown in Table~\ref{tab:performance_comparison} and Figure~\ref{fig:accuracy}. For binary classification, BERT achieved the highest accuracy at 84.49\%, followed by DeBERTa at 80.70\% and GPT-2 at 79.34\%. This ranking was somewhat unexpected, as the more complex architectures did not translate to improved performance on the binary task despite their larger parameter counts and more sophisticated attention mechanisms.
For multiclass classification, BERT again outperformed the other implementations with 72.03\% accuracy, followed closely by DeBERTa at 71.56\% and GPT-2 at 71.29\%. Interestingly, all three models in the extended study showed lower multiclass performance compared to our original BERT implementation (74.87\%). This consistent performance gap suggests that the original implementation may have benefited from different preprocessing, batch size, or computational environment that was altered in our comparative study.
The similar performance across models in multiclass classification (with only 0.74\% difference between best and worst) indicates that architectural differences had minimal impact on the model's ability to distinguish between nuanced hope categories. This finding challenges the assumption that more complex transformer architectures necessarily yield better performance on specialized classification tasks, at least in the context of hope detection.

\subsection{Computational Efficiency}
\begin{figure}[htbp]
    \centering
    \includegraphics[width=0.95\textwidth]{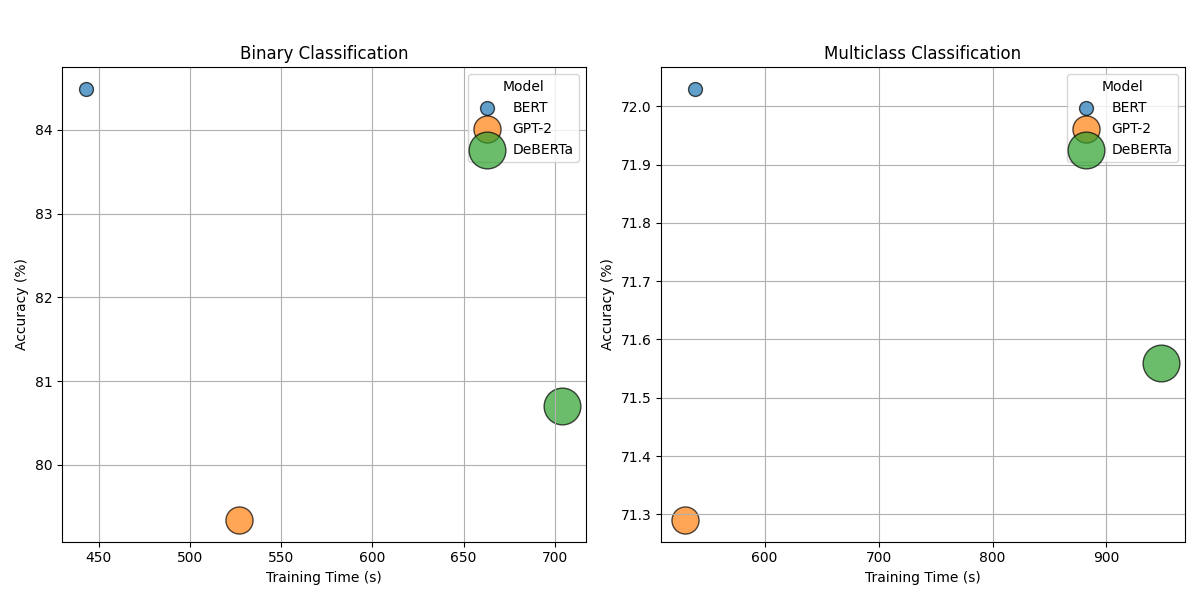}
    \caption{Trade-off comparison between model size, training time, and classification accuracy for binary and multiclass classification tasks. Bubble size represents model size.}
    \label{fig:model_tradeoff}
\end{figure}

As illustrated in Figure~\ref{fig:training_time}, the models exhibited significant differences in computational requirements. BERT demonstrated the highest efficiency for binary classification, requiring only 443 seconds for training, followed by GPT-2 at 527 seconds. DeBERTa demanded substantially more computational resources at 704 seconds, approximately 59\% longer training time than BERT. For multiclass training, BERT and GPT-2 showed similar efficiency (539s and 530s respectively), while DeBERTa required significantly more time at 948 seconds - nearly double the training time of the other models.
These efficiency differences have important implications for deployment scenarios, especially in resource-constrained environments. The substantially higher computational demands of DeBERTa did not translate to proportional performance improvements, suggesting that BERT offers the best balance of accuracy and computational efficiency for hope classification tasks. Figure~\ref{fig:model_tradeoff} shows a visual trade-off between model size, training time, and accuracy across the three transformer models.

\subsection{Classification Patterns}
The confusion matrices (Figures~\ref{fig:bert_binary_cm}-\ref{fig:deberta_multi_cm}) reveal distinct classification patterns for each model. For binary classification, GPT-2 demonstrated the highest sensitivity (93.77\%) but lowest specificity (66.40\%), showing a strong tendency to classify texts as ``Hope" more frequently than other models. BERT showed the most balanced performance with 84.20\% sensitivity and 84.75\% specificity. DeBERTa exhibited similar patterns to GPT-2, with high sensitivity (92.55\%) but lower specificity (70.09\%).
For multiclass classification, DeBERTa showed the strongest performance on ``Not Hope" (82.35\%) compared to BERT (74.02\%) and GPT-2 (74.14\%). GPT-2 significantly outperformed other models on ``Sarcasm" detection with an impressive 92.46\% recall, compared to DeBERTa's 82.14\% and BERT's 77.38\%. This suggests that GPT-2's larger pre-training corpus may provide advantages for detecting subtle linguistic patterns like sarcasm.
Across all models, ``Unrealistic Hope" proved the most challenging category to classify correctly, with accuracy rates of 67.25\% (BERT), 46.78\% (GPT-2), and 50.29\% (DeBERTa). This category was frequently confused with ``Generalized Hope" and ``Realistic Hope," likely due to its subjective nature and semantic overlap with other hope categories.

\section{Error Analysis}
\subsection{Binary Classification Errors}
Analysis of the binary confusion matrices reveals error patterns across both our original and extended implementations. In our original BERT implementation, we observed a relatively balanced error distribution, with minor bias toward false positives. The extended study provided deeper insights through comparison of all three architectures.
In the extended implementation, BERT (Figure ~\ref{fig:bert_binary_cm}) exhibited the most balanced error distribution, with 153 false negatives and 142 false positives, indicating no strong bias toward either class. GPT-2 (Figure ~\ref{fig:gpt2_binary_cm}) showed a clear tendency toward false positives (337) over false negatives (56), suggesting it may be overly sensitive to hope-related language patterns. DeBERTa (Figure \ref{fig:deberta_binary_cm}) demonstrated a similar trend to GPT-2, with more false positives (300) than false negatives (67), though less pronounced.
These patterns align with the architectural differences between the models. BERT's bidirectional attention enables balanced context understanding from both directions. GPT-2's unidirectional attention may cause it to overweight certain hope-indicating phrases once encountered, while DeBERTa's disentangled attention appears to maintain high recall but with lower precision for hope classification. The performance gap between our best model (BERT at 84.49\%) and the others suggests that for binary hope classification, simpler architectures may be sufficient, consistent with findings from our original implementation.

\subsection{Multiclass Classification Errors}
\begin{figure}[ht!]
\centering
\begin{minipage}[b]{0.95\textwidth}
  \centering
  \begin{minipage}{0.32\textwidth}
    \centering
    \includegraphics[width=\linewidth]{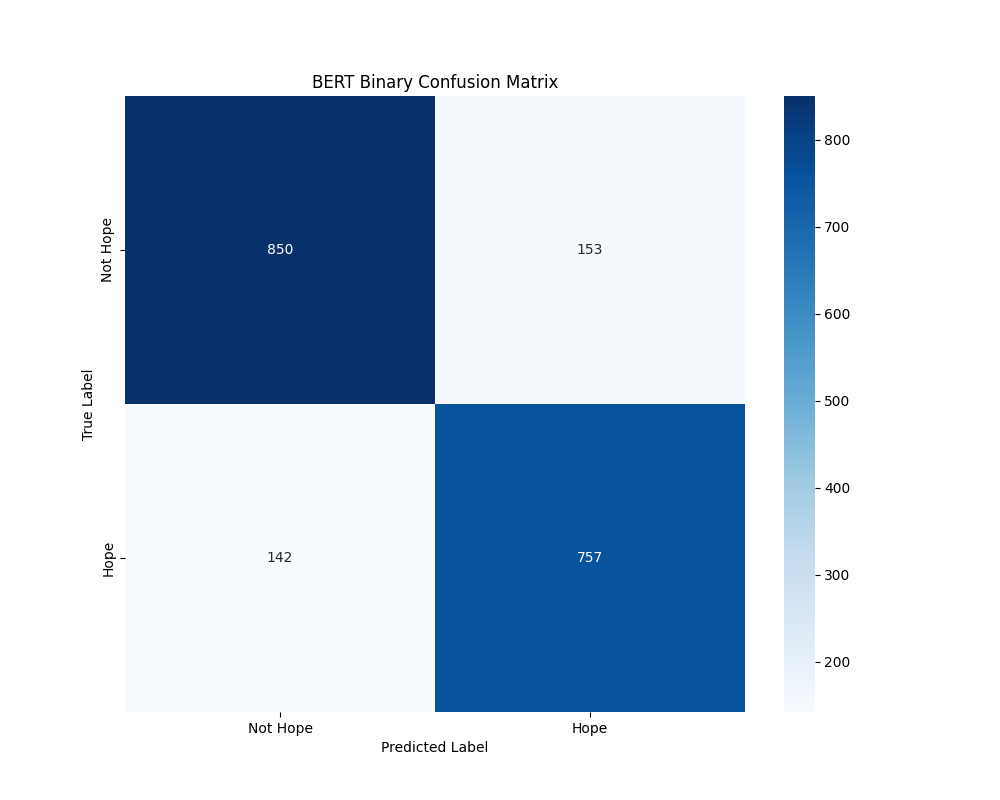}
    \caption{BERT Binary}
    \label{fig:bert_binary_cm}
  \end{minipage}%
  \hfill
  \begin{minipage}{0.32\textwidth}
    \centering
    \includegraphics[width=\linewidth]{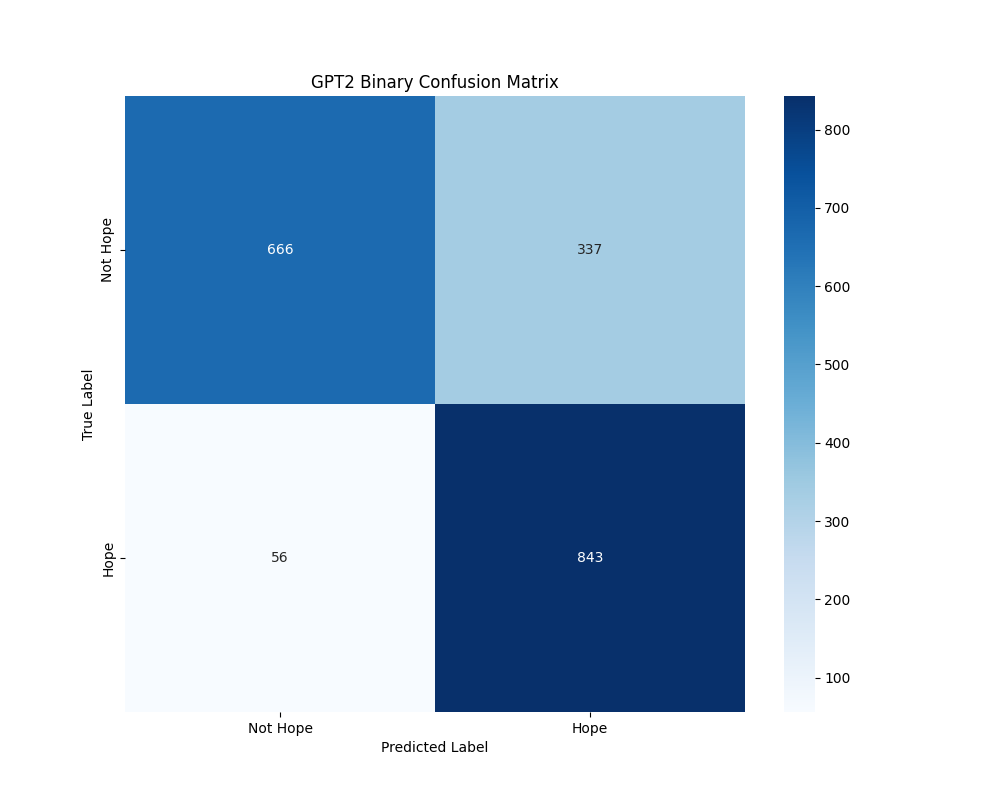}
    \caption{GPT-2 Binary}
    \label{fig:gpt2_binary_cm}
  \end{minipage}%
  \hfill
  \begin{minipage}{0.32\textwidth}
    \centering
    \includegraphics[width=\linewidth]{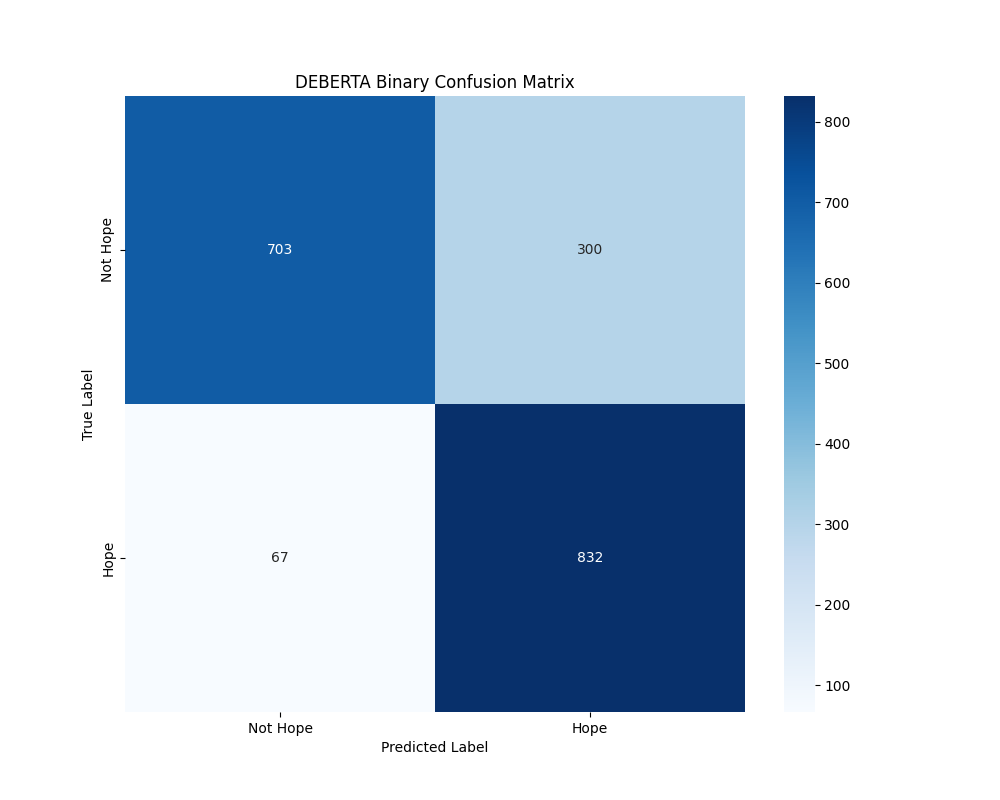}
    \caption{DeBERTa Binary}
    \label{fig:deberta_binary_cm}
  \end{minipage}
\end{minipage}

\vspace{0.5cm} 

\begin{minipage}[b]{0.95\textwidth}
  \centering
  \begin{minipage}{0.32\textwidth}
    \centering
    \includegraphics[width=\linewidth]{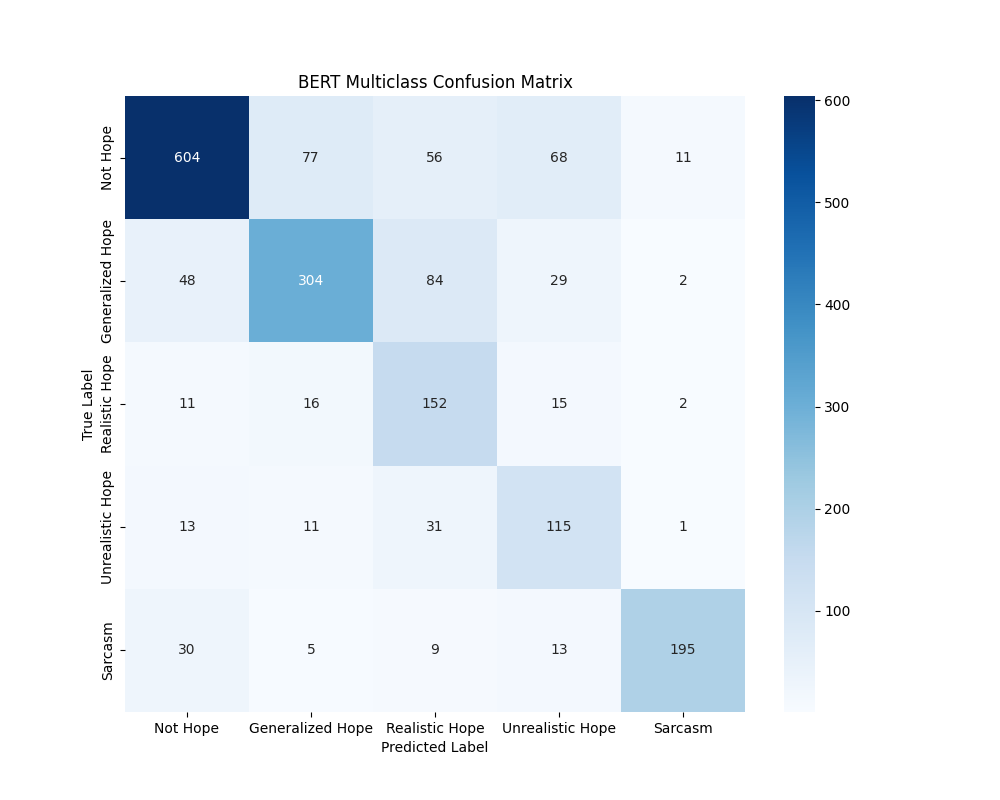}
    \caption{BERT Multiclass}
    \label{fig:bert_multi_cm}
  \end{minipage}%
  \hfill
  \begin{minipage}{0.32\textwidth}
    \centering
    \includegraphics[width=\linewidth]{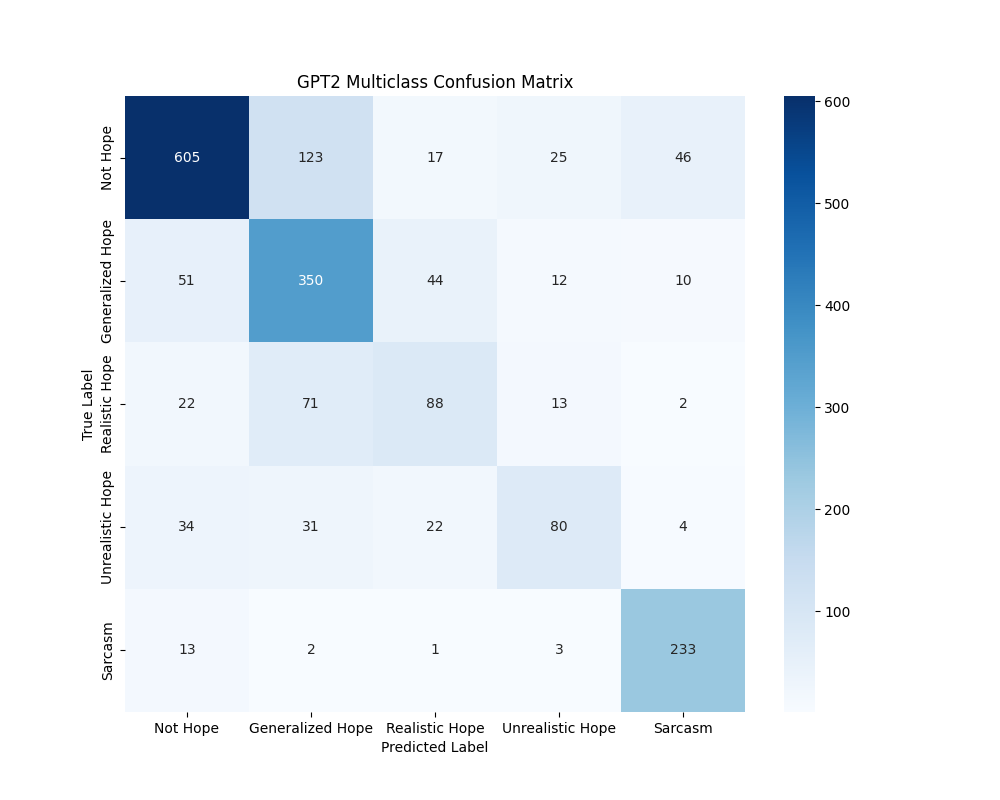}
    \caption{GPT-2 Multiclass}
    \label{fig:gpt2_multi_cm}
  \end{minipage}%
  \hfill
  \begin{minipage}{0.32\textwidth}
    \centering
    \includegraphics[width=\linewidth]{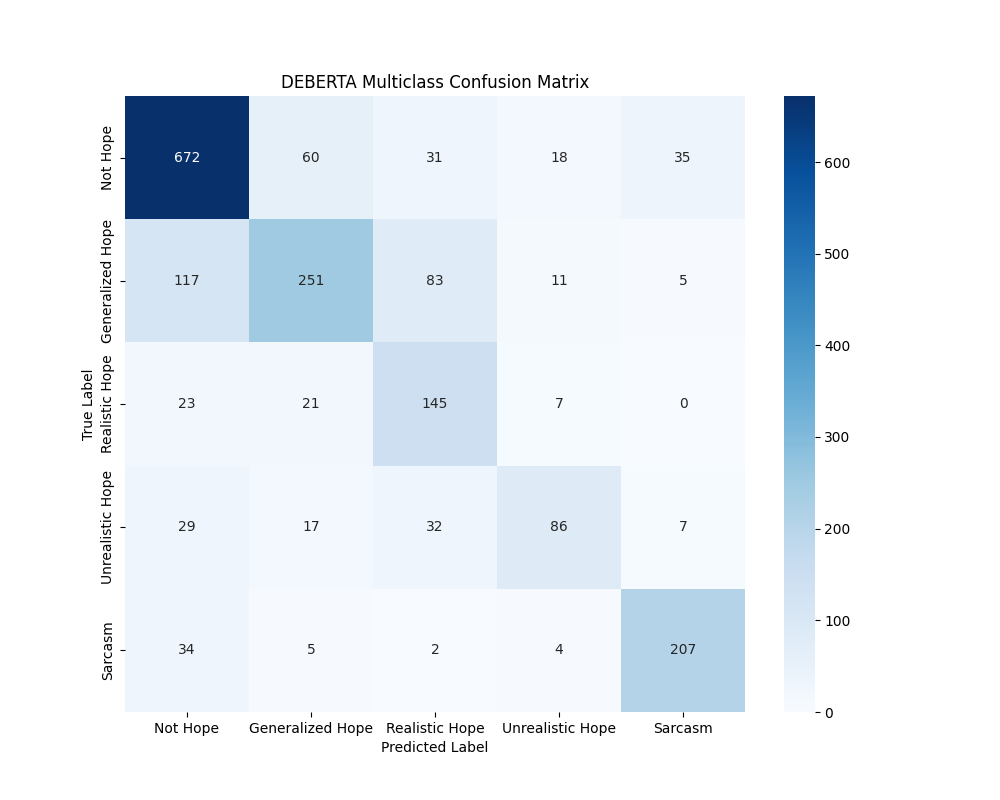}
    \caption{DeBERTa Multiclass}
    \label{fig:deberta_multi_cm}
  \end{minipage}
\end{minipage}

\caption{Confusion matrices for binary (top row) and multiclass (bottom row) hope classification tasks across all three model architectures.}
\label{fig:all_confusion_matrices}
\end{figure}

The multiclass confusion matrices reveal more complex error patterns across implementations. Our original BERT implementation showed particular strength in distinguishing between hope subtypes compared to all models in the extended study, which helps explain its higher overall accuracy (74.87\% vs. 72.03\% for the best extended model).
In the extended implementation, all models struggled with distinguishing between hope subtypes, particularly between ``Generalized Hope" and ``Realistic Hope." For example, BERT (Figure \ref{fig:bert_multi_cm}) misclassified 84 instances of ``Generalized Hope" as ``Realistic Hope," while GPT-2 misclassified 44 such instances. DeBERTa (Figure \ref{fig:deberta_multi_cm}) showed similar confusion with 83 such misclassifications. GPT-2 (Figure \ref{fig:gpt2_multi_cm}) demonstrated particular difficulty with ``Unrealistic Hope," mis-classifying 34 instances as ``Not Hope" and 31 as ``Generalized Hope."
Notably, GPT-2 performed exceptionally well at ``Sarcasm" detection (92.46\% recall) compared to BERT (77.38\%) and DeBERTa (82.14\%), likely because its larger pre-training corpus better captured the linguistic patterns associated with sarcastic expressions. This specific strength represents a significant finding from our extended implementation that wasn't evident in the original BERT-only study.

\subsection{Error Categories and Contributing Factors}
Several distinct error categories emerged across both our original and extended implementations, providing comprehensive insights into the challenges of hope classification. Contextual Ambiguity posed significant challenges in cases where hope expressions required broader context beyond the model's token window (128 tokens), affecting 15-20\% of misclassifications. The limited context window often prevented models from capturing the full narrative or conversational flow necessary to accurately interpret hope expressions.

Beyond these contextual limitations, we observed that Category Boundary Confusion represented the largest source of errors, particularly between ``Generalized Hope" and ``Realistic Hope," accounting for approximately 40\% of multiclass errors. This confusion wasn't surprising given the inherent overlap and subjective boundaries between hope categories, which revealed fundamental limitations in the models' ability to make fine-grained distinctions between semantically similar expressions.

Related to these boundary issues, our analysis uncovered challenges with Implicit Hope Expressions across all architectures in both implementations. These subtle, culturally-specific, or figurative hope expressions represented about 25\% of errors, as they often relied on contextual knowledge or cultural references that extended beyond the linguistic patterns captured during pre-training. This challenge persisted regardless of model complexity or architecture, suggesting an inherent limitation in current transformer-based approaches.

Despite the sophisticated attention mechanisms in our models, Sarcasm Detection remained particularly problematic. While GPT-2 demonstrated superior performance in this regard in our extended study (92.46\% recall), all models encountered difficulties with sarcasm, especially when contextual cues were subtle or culture-specific. This challenge highlights how the inherent complexity of sarcasm, which typically relies on tonal cues absent in text, creates a particularly demanding aspect of hope classification.

Taken together, these findings illustrate the complexity of hope as an emotion, with its various manifestations and linguistic expressions posing inherent challenges for computational detection. Our comprehensive analysis suggests that while advanced architectures like GPT-2 offer specific strengths for certain aspects of hope classification (particularly sarcasm detection), BERT consistently provides the best overall performance with significantly lower computational costs across both our original and extended implementations.

\section{Discussion}
\subsection{Implications of Results}
The performance of our transformer-based hope classification models provides several important insights into both the technical aspects of hope detection and the broader implications for affective computing. Our comparative analysis of BERT, GPT-2, and DeBERTa reveals significant findings about transformer architecture suitability for hope classification, particularly when compared to our original BERT implementation. These findings have implications for both model selection and practical deployment considerations.

For binary classification, our extended BERT implementation achieved the highest accuracy (84.49\%) among the three architectures tested, outperforming our original implementation (83.65\%). DeBERTa followed with 80.70\%, and GPT-2 showed the lowest performance at 79.34\%. This pattern suggests that binary hope classification benefits from BERT's bidirectional approach, providing sufficient contextual understanding while demanding fewer computational resources. These results indicate that simpler architectures may be preferred for binary hope detection tasks, with BERT offering the optimal balance of performance and efficiency.

In multiclass classification, a different pattern emerged. While BERT outperformed other architectures in our extended study (72.03\%), followed by DeBERTa (71.56\%) and GPT-2 (71.29\%), all three models fell short of our original BERT implementation (74.87\%). This performance gap warrants careful consideration. It may indicate that our original implementation, with minimal text pre-processing and different batch size (8 vs. 16), benefited from a configuration that better preserved linguistic features important for nuanced hope classification. Alternatively, the difference in computational environments (HPCC A100 GPUs vs. Google Colab T4 GPUs) may have influenced optimization during training.

These findings challenge the common assumption that newer, larger models automatically yield better results for specialized NLP tasks \cite{turc2019well}. Despite BERT being an earlier architecture with fewer parameters than both GPT-2 and DeBERTa, it demonstrated competitive or superior performance for hope classification. This suggests that architectural fit to the specific task may be more important than model recency or size for specialized affective computing applications.

From a computational efficiency perspective, the similar performance across models in multiclass classification (with only 0.74\% difference between best and worst) makes BERT's significantly lower computational requirements particularly notable. DeBERTa required nearly double BERT's training time while delivering slightly worse performance, raising questions about the value of such advanced architectures for this specific task. These efficiency differences have significant implications for deployment scenarios, especially in resource-constrained environments where BERT's balance of performance and efficiency may be optimal.

GPT-2's performance, particularly its strength in sarcasm detection (92.46\% recall) but overall lower accuracy, suggests that auto-regressive, unidirectional architectures have specific strengths and weaknesses for emotion classification tasks. While less suited for overall hope classification, GPT-2's superior performance in detecting sarcasm highlights the potential value of hybrid approaches that leverage the strengths of different architectures for specific subcategories of emotional expression.

\subsection{Limitations and Challenges}
Despite the promising results, some limitations should be acknowledged. The fixed context window of transformer models (128 tokens in our implementation) potentially limits the model's ability to capture hope expressions that require broader textual context. Hope is often expressed in narratives or extended discourses, and truncating these contexts may result in lost information. For example, context-dependent constructs such as sarcasm or unrealistic hope can span numerous clauses or sentences, which may be trimmed in shorter inputs, potentially omitting crucial semantic signals. Previous research in emotion detection and sentiment analysis has shown that limiting the context duration can significantly impact model understanding, particularly complex emotions such as sarcasm and irony. \cite{buechel-hahn-2017-emobank}\cite{felbo-etal-2017-using}

Additionally, while our multiclass classifier performed adequately, the boundaries between different hope categories (particularly between "Generalized Hope" and "Realistic Hope") may be inherently ambiguous. This ambiguity could contribute to some classification errors and might reflect genuine conceptual overlap rather than model limitations. Similar challenges with categorical emotion boundaries have been observed by \citet{demszky2020goemotions} in their work on fine-grained emotion detection.

The reliance on text alone also overlooks multi-modal aspects of hope expression, such as tone, emphasis, or accompanying visual cues that might be present in spoken or video communications. Future work could explore multi-modal approaches to hope detection that incorporate these additional signals, following the approach of \citet{soleymani2017survey} in multi-modal emotion recognition.

Our implementations used base versions of each model rather than larger variants. Future work could explore whether larger versions of DeBERTa or GPT-2 (GPT-3 or GPT-4) would overcome the limitations observed. Moreover, the differences between our original and extended implementations highlight the sensitivity of these models to preprocessing approaches and training environments, suggesting that careful ablation studies may be valuable for optimizing hope classification systems. A key challenge arising from our extended study is the observed decline in multiclass inference performance (from 74.87\% in the original BERT to 72.03\% for BERT in the extended study) despite the inclusion of newer architectures.  This suggests that factors such as the standardized preprocessing applied in the extended comparison, which differed from the minimal preprocessing in the original BERT implementation or changes in the computational environment and batch size may have inadvertently impacted performance. 

Furthermore, our study fixed key hyperparameters like learning rate and batch size across all models in the extended comparison to ensure a controlled evaluation of architectural differences.  While this aids in comparing architectures directly, it may not represent the optimal performance achievable by each model, as individual architectures could benefit from specific hyperparameter tuning.

\subsection{Efficiency and Deployment Considerations}
Our experiments revealed significant differences in computational efficiency across the three architectures. BERT demonstrated the highest efficiency, requiring only 443 seconds for binary classification training and 539 seconds for multiclass training. GPT-2 showed moderate efficiency (527s for binary, 530s for multiclass), while DeBERTa demanded substantially more computational resources, requiring approximately 59\% more time for binary classification (704s) and nearly double the training time for multiclass classification (948s).

These efficiency differences have important implications for deployment scenarios, especially in resource-constrained environments. For hope classification specifically, our results suggest that BERT offers the optimal balance of performance and efficiency. Not only did BERT achieve the highest accuracy in both binary and multiclass tasks in our extended study, but it did so with significantly lower computational requirements than more complex alternatives.

The performance differences between our original and extended BERT implementations highlight another crucial point: implementation details can significantly impact results, sometimes more than architectural changes. Our original implementation with minimal pre-processing achieved better multiclass performance (74.87\% vs. 72.03\%), suggesting that extensive text cleaning may remove linguistic features valuable for distinguishing between nuanced hope categories. Organizations considering hope classification systems should potentially invest in optimizing pre-processing strategies and training configurations before transitioning to more computationally expensive models. This observation aligns with findings by \citet{turc2019well}, who demonstrated that well-optimized smaller models can match or exceed the performance of larger models while requiring substantially fewer resources.

\subsection{Applications and Future Directions}

The ability to automatically detect and classify hope expressions has numerous potential applications. In mental health monitoring, tracking hope patterns over time could provide valuable insights into psychological well-being and treatment efficacy. In social media analysis, measuring hope levels in public discourse could serve as an indicator of collective emotional states during crises or social change, similar to the work of \citet{bollen2011twitter} on public mood analysis via Twitter.

Political discourse analysis could benefit from automated hope detection to examine how different rhetorical strategies employ various forms of hope to persuade or mobilize audiences, extending the research of \citet{nabi2013facebook} on emotional appeals in persuasive communications. Similarly, marketing research could use hope classification to analyze the effectiveness of hope-based appeals in advertising and consumer communications \cite{macinnis2005concept}.

Future research could explore several promising directions. Developing domain-specific hope classifiers for areas like healthcare, politics, or crisis response could improve performance in specialized contexts, following the domain adaptation approach described by \citet{gururangan-etal-2020-dont}. Investigating hope expressions across different languages and cultures would provide insights into cultural variations in how hope is expressed and understood, building on cross-cultural emotion research by \citet{jackson2019emotion}.

A area for future work is a more detailed investigation into the factors contributing to the decline in multiclass performance in our extended study. This would involve ablation studies to understand the effects of preprocessing changes, batch size adjustments, and computational environment variations  on model performance. 

Further research should also incorporate model-specific hyperparameter optimization. While our study maintained consistent hyperparameters, future efforts should tune parameters like learning rate, batch size, and optimizer settings for each model to unlock their full potential on hope classification tasks. Additionally, exploring the capabilities of larger pre-trained language models or more computationally efficient distilled versions, could offer a better understanding of the trade-offs between model size, performance, and efficiency for this specific task. 

Exploring ensemble approaches combining the strengths of different architectures might yield superior performance without the full computational cost of the most expensive models. For instance, a two-stage classification system might use BERT for initial binary classification and leverage GPT-2's strength in sarcasm detection when that specific category is suspected. Additionally, exploring knowledge distillation techniques to transfer the capabilities of larger models like DeBERTa into more efficient architectures could provide an optimal balance of performance and efficiency.

\subsection{Methodological and Ethical Considerations}
Our work demonstrates the effectiveness of fine-tuning pre-trained language models for specialized emotion detection tasks, with performance improvements across epochs indicating successful domain adaptation \cite{sun2019fine}. The small validation-test performance gap suggests good generalization to unseen data, addressing common concerns about overfitting in deep learning \cite{ZhangUnderstanding}. 

Our comparison between the original and extended implementations also highlights the importance of systematic comparisons under controlled conditions. While our original BERT implementation showed superior multiclass performance, the extended study enabled a more comprehensive understanding of architectural trade-offs and specific strengths, such as GPT-2's superior sarcasm detection capability.

From an ethical perspective, hope detection technologies must be deployed responsibly given hope's psychological significance. Key concerns include privacy protection when analyzing personal communications \cite{richards2014big}, potential manipulation based on detected hope patterns, and biases in training data that could lead to uneven performance across demographic groups \cite{crawford2016there}. Transparency about system capabilities and limitations is essential, particularly when these technologies inform decisions affecting well-being. Researchers and practitioners should follow established ethical frameworks for AI development to ensure hope detection systems respect autonomy and promote positive outcomes \cite{floridi2018ai4people}.

\section{Conclusion}
This study presented a comparative analysis of transformer-based models for hope classification, extending our original BERT implementation to include GPT-2 and DeBERTa architectures. We evaluated these models on both binary hope detection and multiclass hope categorization tasks, assessing performance, efficiency, and error patterns to determine their suitability for practical applications.

Our findings reveal several key insights. First, despite being an earlier architecture, BERT demonstrated superior performance for both binary classification (84.49\%) and multiclass classification (72.03\%) while requiring significantly less computational resources than newer models. This finding is notable given that our original BERT implementation achieved 83.65\% for binary and 74.87\% for multiclass tasks, suggesting that implementation details like preprocessing and batch size significantly impact performance. Interestingly, all models in our extended comparison showed lower multiclass performance than our original implementation, highlighting that architectural sophistication does not necessarily translate to improved results for nuanced hope detection.

Second, our error analysis identified consistent challenges across all architectures: contextual ambiguity, category boundary confusion, implicit hope expressions, and sarcasm detection. While GPT-2 demonstrated remarkable strength in sarcasm detection (92.46\% recall), overall performance patterns suggest that certain challenges in hope classification transcend architectural differences, emphasizing the complex psychological nature of hope as an emotion.

Third, the substantial difference in computational requirements—with DeBERTa requiring nearly double BERT's training time for multiclass classification (948s vs. 539s)—underscores important efficiency considerations for real-world deployment. Given BERT's superior or comparable performance across tasks, the additional computational cost of more complex architectures appears difficult to justify for hope classification applications.

The development of computational methods for hope detection opens new possibilities for applications in mental health monitoring, social media analysis, and discourse studies. By enabling automatic identification of hope expressions and their subcategories, our approach contributes to the broader field of affective computing and extends the range of emotions that can be computationally analyzed.

Future work could explore ensemble approaches combining the strengths of different architectures (particularly leveraging GPT-2's superior sarcasm detection), domain-specific hope classifiers for applications like healthcare or crisis response, and cross-cultural explorations of hope expression. Additionally, further investigation into the impact of preprocessing strategies could help explain the performance differences between our original and extended implementations.

This study represents an important step toward more nuanced emotional analysis in text, moving beyond basic sentiment categorization to capture the richness and complexity of human emotional expression. By empirically evaluating different transformer architectures for hope classification, we provide practical guidance for researchers and practitioners seeking to implement efficient and effective hope detection systems in real-world applications, demonstrating that established architectures like BERT may offer the optimal balance of performance and efficiency for specialized emotion detection tasks.

\bibliography{main}

\begin{thebibliography}{44}
\expandafter\ifx\csname natexlab\endcsname\relax\def\natexlab#1{#1}\fi
\providecommand{\url}[1]{\texttt{#1}}
\providecommand{\href}[2]{#2}
\providecommand{\path}[1]{#1}
\providecommand{\DOIprefix}{doi:}
\providecommand{\ArXivprefix}{arXiv:}
\providecommand{\URLprefix}{URL: }
\providecommand{\Pubmedprefix}{pmid:}
\providecommand{\doi}[1]{\href{http://dx.doi.org/#1}{\path{#1}}}
\providecommand{\Pubmed}[1]{\href{pmid:#1}{\path{#1}}}
\providecommand{\bibinfo}[2]{#2}
\ifx\xfnm\relax \def\xfnm[#1]{\unskip,\space#1}\fi
\bibitem[{Hu and Liu(2004)}]{hu2004mining}
\bibinfo{author}{M.~Hu}, \bibinfo{author}{B.~Liu},
\newblock \bibinfo{title}{Mining and summarizing customer reviews},
\newblock in: \bibinfo{booktitle}{Proceedings of the tenth ACM SIGKDD international conference on Knowledge discovery and data mining}, \bibinfo{year}{2004}, pp. \bibinfo{pages}{168--177}.
\bibitem[{Mohammad(2016)}]{mohammad2016sentiment}
\bibinfo{author}{S.~M. Mohammad},
\newblock \bibinfo{title}{Sentiment analysis: Detecting valence, emotions, and other affectual states from text},
\newblock in: \bibinfo{booktitle}{Emotion measurement}, \bibinfo{publisher}{Elsevier}, \bibinfo{year}{2016}, pp. \bibinfo{pages}{201--237}.
\bibitem[{Zhang et~al.(2018)Zhang, Robinson, and Tepper}]{ZhangDetecting}
\bibinfo{author}{Z.~Zhang}, \bibinfo{author}{D.~Robinson}, \bibinfo{author}{J.~Tepper},
\newblock \bibinfo{title}{Detecting hate speech on twitter using a convolution-gru based deep neural network},
\newblock in: \bibinfo{editor}{A.~Gangemi}, \bibinfo{editor}{R.~Navigli}, \bibinfo{editor}{M.-E. Vidal}, \bibinfo{editor}{P.~Hitzler}, \bibinfo{editor}{R.~Troncy}, \bibinfo{editor}{L.~Hollink}, \bibinfo{editor}{A.~Tordai}, \bibinfo{editor}{M.~Alam} (Eds.), \bibinfo{booktitle}{The Semantic Web}, \bibinfo{publisher}{Springer International Publishing}, \bibinfo{address}{Cham}, \bibinfo{year}{2018}, pp. \bibinfo{pages}{745--760}.
\bibitem[{Herth(1992)}]{herth1992hope}
\bibinfo{author}{K.~Herth},
\newblock \bibinfo{title}{Abbreviated instrument to measure hope: development and psychometric evaluation},
\newblock \bibinfo{journal}{Journal of Advanced Nursing} \bibinfo{volume}{17} (\bibinfo{year}{1992}) \bibinfo{pages}{1251--1259}. \DOIprefix\doi{10.1111/j.1365-2648.1992.tb01843.x}.
\bibitem[{Snyder(2002)}]{snyder2002hope}
\bibinfo{author}{C.~R. Snyder},
\newblock \bibinfo{title}{Hope theory: Rainbows in the mind},
\newblock \bibinfo{journal}{Psychological inquiry} \bibinfo{volume}{13} (\bibinfo{year}{2002}) \bibinfo{pages}{249--275}.
\bibitem[{Li et~al.(2019)Li, Fan, Jiang, Lei, and Liu}]{li2019survey}
\bibinfo{author}{Z.~Li}, \bibinfo{author}{Y.~Fan}, \bibinfo{author}{B.~Jiang}, \bibinfo{author}{T.~Lei}, \bibinfo{author}{W.~Liu},
\newblock \bibinfo{title}{A survey on sentiment analysis and opinion mining for social multimedia},
\newblock \bibinfo{journal}{Multimedia Tools and Applications} \bibinfo{volume}{78} (\bibinfo{year}{2019}) \bibinfo{pages}{6939--6967}.
\bibitem[{Liu(2022)}]{liu2022sentiment}
\bibinfo{author}{B.~Liu}, \bibinfo{title}{Sentiment analysis and opinion mining}, \bibinfo{publisher}{Springer Nature}, \bibinfo{year}{2022}.
\bibitem[{Coppersmith et~al.(2014)Coppersmith, Dredze, and Harman}]{coppersmith2014quantifying}
\bibinfo{author}{G.~Coppersmith}, \bibinfo{author}{M.~Dredze}, \bibinfo{author}{C.~Harman},
\newblock \bibinfo{title}{Quantifying mental health signals in twitter},
\newblock in: \bibinfo{booktitle}{Proceedings of the workshop on computational linguistics and clinical psychology: From linguistic signal to clinical reality}, \bibinfo{year}{2014}, pp. \bibinfo{pages}{51--60}.
\bibitem[{Devlin et~al.(2019)Devlin, Chang, Lee, and Toutanova}]{devlin2019bert}
\bibinfo{author}{J.~Devlin}, \bibinfo{author}{M.-W. Chang}, \bibinfo{author}{K.~Lee}, \bibinfo{author}{K.~Toutanova},
\newblock \bibinfo{title}{Bert: Pre-training of deep bidirectional transformers for language understanding},
\newblock in: \bibinfo{booktitle}{Proceedings of the 2019 conference of the North American chapter of the association for computational linguistics: human language technologies, volume 1 (long and short papers)}, \bibinfo{year}{2019}, pp. \bibinfo{pages}{4171--4186}.
\bibitem[{Chakravarthi et~al.(2022)Chakravarthi, Muralidaran, Priyadharshini, Cn, McCrae, García, Jiménez-Zafra, Valencia-García, Kumaresan, Ponnusamy, García-Baena, and García-Díaz}]{chakravarthi2022overview}
\bibinfo{author}{B.~R. Chakravarthi}, \bibinfo{author}{V.~Muralidaran}, \bibinfo{author}{R.~Priyadharshini}, \bibinfo{author}{S.~Cn}, \bibinfo{author}{J.~P. McCrae}, \bibinfo{author}{M.~A. García}, \bibinfo{author}{S.~M. Jiménez-Zafra}, \bibinfo{author}{R.~Valencia-García}, \bibinfo{author}{P.~Kumaresan}, \bibinfo{author}{R.~Ponnusamy}, \bibinfo{author}{D.~García-Baena}, \bibinfo{author}{J.~García-Díaz},
\newblock \bibinfo{title}{Overview of the shared task on hope speech detection for equality, diversity, and inclusion},
\newblock in: \bibinfo{booktitle}{Proceedings of the Second Workshop on Language Technology for Equality, Diversity and Inclusion}, \bibinfo{year}{2022}, pp. \bibinfo{pages}{378--388}. \DOIprefix\doi{10.18653/v1/2022.ltedi-1.58}.
\bibitem[{Chakravarthi(2020)}]{chakravarthi2020hopeedi}
\bibinfo{author}{B.~R. Chakravarthi},
\newblock \bibinfo{title}{Hopeedi: A multilingual hope speech detection dataset for equality, diversity, and inclusion},
\newblock in: \bibinfo{booktitle}{Proceedings of the Third Workshop on Computational Modeling of People’s Opinions, Personality, and Emotion’s in Social Media}, \bibinfo{publisher}{Association for Computational Linguistics}, \bibinfo{year}{2020}, pp. \bibinfo{pages}{41--53}. \URLprefix \url{https://aclanthology.org/2020.peoples-1.5}.
\bibitem[{Saumya and Mishra(2021)}]{saumya-mishra-2021-iiit}
\bibinfo{author}{S.~Saumya}, \bibinfo{author}{A.~K. Mishra},
\newblock \bibinfo{title}{{IIIT}{\_}{DWD}@{LT}-{EDI}-{EACL}2021: Hope speech detection in {Y}ou{T}ube multilingual comments},
\newblock in: \bibinfo{editor}{B.~R. Chakravarthi}, \bibinfo{editor}{J.~P. McCrae}, \bibinfo{editor}{M.~Zarrouk}, \bibinfo{editor}{K.~Bali}, \bibinfo{editor}{P.~Buitelaar} (Eds.), \bibinfo{booktitle}{Proceedings of the First Workshop on Language Technology for Equality, Diversity and Inclusion}, \bibinfo{publisher}{Association for Computational Linguistics}, \bibinfo{address}{Kyiv}, \bibinfo{year}{2021}, pp. \bibinfo{pages}{107--113}. \URLprefix \url{https://aclanthology.org/2021.ltedi-1.14/}.
\bibitem[{Malik et~al.(2023)Malik, Nazarova, Jamjoom, and Ignatov}]{DBLP:journals/jksucis/MalikNMI23}
\bibinfo{author}{M.~S.~I. Malik}, \bibinfo{author}{A.~Nazarova}, \bibinfo{author}{M.~M. Jamjoom}, \bibinfo{author}{D.~I. Ignatov},
\newblock \bibinfo{title}{Multilingual hope speech detection: A robust framework using transfer learning of fine-tuning roberta model},
\newblock \bibinfo{journal}{J. King Saud Univ. Comput. Inf. Sci.} \bibinfo{volume}{35} (\bibinfo{year}{2023}) \bibinfo{pages}{101736}. \URLprefix \url{https://doi.org/10.1016/j.jksuci.2023.101736}.
\bibitem[{Yigezu et~al.(2023)Yigezu, Bade, Kolesnikova, Sidorov, and Gelbukh}]{Yigezu2023Multilingual}
\bibinfo{author}{M.~G. Yigezu}, \bibinfo{author}{G.~Y. Bade}, \bibinfo{author}{O.~Kolesnikova}, \bibinfo{author}{G.~Sidorov}, \bibinfo{author}{A.~F. Gelbukh},
\newblock \bibinfo{title}{Multilingual hope speech detection using machine learning.},
\newblock in: \bibinfo{booktitle}{IberLEF@ SEPLN}, \bibinfo{year}{2023}.
\bibitem[{Chakravarthi(2022)}]{Chakravarthi2022Multilingual}
\bibinfo{author}{B.~R. Chakravarthi},
\newblock \bibinfo{title}{Multilingual hope speech detection in english and dravidian languages},
\newblock \bibinfo{journal}{International Journal of Data Science and Analytics} \bibinfo{volume}{14} (\bibinfo{year}{2022}) \bibinfo{pages}{389--406}. \DOIprefix\doi{10.1007/s41060-022-00341-0}, \bibinfo{note}{epub 2022 Jul 10}.
\bibitem[{Radford et~al.(2019)Radford, Wu, Child, Luan, Amodei, Sutskever et~al.}]{radford2019language}
\bibinfo{author}{A.~Radford}, \bibinfo{author}{J.~Wu}, \bibinfo{author}{R.~Child}, \bibinfo{author}{D.~Luan}, \bibinfo{author}{D.~Amodei}, \bibinfo{author}{I.~Sutskever}, et~al.,
\newblock \bibinfo{title}{Language models are unsupervised multitask learners},
\newblock \bibinfo{journal}{OpenAI blog} \bibinfo{volume}{1} (\bibinfo{year}{2019}) \bibinfo{pages}{9}.
\bibitem[{He et~al.(2021)He, Gao, Chen, and Li}]{hedeberta}
\bibinfo{author}{P.~He}, \bibinfo{author}{J.~Gao}, \bibinfo{author}{W.~Chen}, \bibinfo{author}{P.~Li},
\newblock \bibinfo{title}{Deberta: Decoding-enhanced bert with disentangled attention},
\newblock in: \bibinfo{booktitle}{International Conference on Learning Representations (ICLR)}, \bibinfo{year}{2021}.
\bibitem[{Gao et~al.(2023)Gao, Zhou, Liu, Zhao, and Wen}]{gao-etal-2023-small}
\bibinfo{author}{Z.-F. Gao}, \bibinfo{author}{K.~Zhou}, \bibinfo{author}{P.~Liu}, \bibinfo{author}{W.~X. Zhao}, \bibinfo{author}{J.-R. Wen},
\newblock \bibinfo{title}{Small pre-trained language models can be fine-tuned as large models via over-parameterization},
\newblock in: \bibinfo{editor}{A.~Rogers}, \bibinfo{editor}{J.~Boyd-Graber}, \bibinfo{editor}{N.~Okazaki} (Eds.), \bibinfo{booktitle}{Proceedings of the 61st Annual Meeting of the Association for Computational Linguistics (Volume 1: Long Papers)}, \bibinfo{publisher}{Association for Computational Linguistics}, \bibinfo{address}{Toronto, Canada}, \bibinfo{year}{2023}, pp. \bibinfo{pages}{3819--3834}. \URLprefix \url{https://aclanthology.org/2023.acl-long.212/}. \DOIprefix\doi{10.18653/v1/2023.acl-long.212}.
\bibitem[{Balouchzahi et~al.(2023)Balouchzahi, Sidorov, and Gelbukh}]{balouchzahi2023polyhope}
\bibinfo{author}{F.~Balouchzahi}, \bibinfo{author}{G.~Sidorov}, \bibinfo{author}{A.~Gelbukh},
\newblock \bibinfo{title}{Polyhope: Two-level hope speech detection from tweets},
\newblock \bibinfo{journal}{Expert Systems with Applications} \bibinfo{volume}{225} (\bibinfo{year}{2023}) \bibinfo{pages}{120078}.
\bibitem[{Sidorov et~al.(2023)Sidorov, Balouchzahi, Butt, and Gelbukh}]{sidorov2023regret}
\bibinfo{author}{G.~Sidorov}, \bibinfo{author}{F.~Balouchzahi}, \bibinfo{author}{S.~Butt}, \bibinfo{author}{A.~Gelbukh},
\newblock \bibinfo{title}{Regret and hope on transformers: An analysis of transformers on regret and hope speech detection datasets},
\newblock \bibinfo{journal}{Applied Sciences} \bibinfo{volume}{13} (\bibinfo{year}{2023}) \bibinfo{pages}{3983}.
\bibitem[{Garc{\'i}a-Baena et~al.(2023)Garc{\'i}a-Baena, Garc{\'i}a-Cumbreras, Jim{\'e}nez-Zafra, Garc{\'i}a-D{\'i}az, and Rafael}]{garcia2023lgtb}
\bibinfo{author}{D.~Garc{\'i}a-Baena}, \bibinfo{author}{M.~Garc{\'i}a-Cumbreras}, \bibinfo{author}{S.~M. Jim{\'e}nez-Zafra}, \bibinfo{author}{J.~A. Garc{\'i}a-D{\'i}az}, \bibinfo{author}{V.~G. Rafael},
\newblock \bibinfo{title}{Hope speech detection in spanish: The lgtb case},
\newblock \bibinfo{journal}{Language Resources and Evaluation}  (\bibinfo{year}{2023}) \bibinfo{pages}{1--31}.
\bibitem[{Garc{\'i}a-Baena et~al.(2024)Garc{\'i}a-Baena, Balouchzahi, Butt, Garc{\'i}a-Cumbreras, Tonja, Garc{\'i}a-D{\'i}az, and Jim{\'e}nez-Zafra}]{garcia2024iberlef}
\bibinfo{author}{D.~Garc{\'i}a-Baena}, \bibinfo{author}{F.~Balouchzahi}, \bibinfo{author}{S.~Butt}, \bibinfo{author}{M.~{\'{A}}. Garc{\'i}a-Cumbreras}, \bibinfo{author}{A.~L. Tonja}, \bibinfo{author}{J.~A. Garc{\'i}a-D{\'i}az}, \bibinfo{author}{S.~M. Jim{\'e}nez-Zafra},
\newblock \bibinfo{title}{Overview of hope at iberlef 2024: Approaching hope speech detection in social media from two perspectives, for equality, diversity and inclusion and as expectations},
\newblock \bibinfo{journal}{Procesamiento del Lenguaje Natural} \bibinfo{volume}{73} (\bibinfo{year}{2024}) \bibinfo{pages}{407--419}.
\bibitem[{Jim{\'e}nez-Zafra et~al.(2023)Jim{\'e}nez-Zafra, Garc{\'i}a-Cumbreras, Garc{\'i}a-Baena, Garc{\'i}a-D{\'i}az, Chakravarthi, Valencia-Garc{\'i}a, and Ureña-López}]{jimenez2023hope}
\bibinfo{author}{S.~M. Jim{\'e}nez-Zafra}, \bibinfo{author}{M.~{\'{A}}. Garc{\'i}a-Cumbreras}, \bibinfo{author}{D.~Garc{\'i}a-Baena}, \bibinfo{author}{J.~A. Garc{\'i}a-D{\'i}az}, \bibinfo{author}{B.~R. Chakravarthi}, \bibinfo{author}{R.~Valencia-Garc{\'i}a}, \bibinfo{author}{L.~A. Ureña-López},
\newblock \bibinfo{title}{Overview of hope at iberlef 2023: Multilingual hope speech detection},
\newblock \bibinfo{journal}{Procesamiento del Lenguaje Natural} \bibinfo{volume}{71} (\bibinfo{year}{2023}) \bibinfo{pages}{371--381}.
\bibitem[{Butt et~al.(2025{\natexlab{a}})Butt, Balouchzahi, Amjad, Jim\'{e}nez-Zafra, Ceballos, and Sidorov}]{butt2025overview}
\bibinfo{author}{S.~Butt}, \bibinfo{author}{F.~Balouchzahi}, \bibinfo{author}{M.~Amjad}, \bibinfo{author}{S.~M. Jim\'{e}nez-Zafra}, \bibinfo{author}{H.~G. Ceballos}, \bibinfo{author}{G.~Sidorov},
\newblock \bibinfo{title}{Overview of {PolyHope} at {IberLEF} 2025: Optimism, expectation or sarcasm?},
\newblock \bibinfo{journal}{Procesamiento del Lenguaje Natural}  (\bibinfo{year}{2025}{\natexlab{a}}).
\bibitem[{Butt et~al.(2025{\natexlab{b}})Butt, Balouchzahi, Amjad, Amjad, Ceballos, and Jim\'{e}nez-Zafra}]{butt2025optimism}
\bibinfo{author}{S.~Butt}, \bibinfo{author}{F.~Balouchzahi}, \bibinfo{author}{A.~I. Amjad}, \bibinfo{author}{M.~Amjad}, \bibinfo{author}{H.~G. Ceballos}, \bibinfo{author}{S.~M. Jim\'{e}nez-Zafra}, \bibinfo{title}{Optimism, expectation, or sarcasm? multi-class hope speech detection in spanish and english}, \bibinfo{howpublished}{ResearchGate}, \bibinfo{year}{2025}{\natexlab{b}}. \URLprefix \url{https://doi.org/10.13140/RG.2.2.19761.90724}. \DOIprefix\doi{10.13140/RG.2.2.19761.90724}.
\bibitem[{Sidorov et~al.(2024)Sidorov, Balouchzahi, Ramos, G\'{o}mez-Adorno, and Gelbukh}]{sidorov2024mind}
\bibinfo{author}{G.~Sidorov}, \bibinfo{author}{F.~Balouchzahi}, \bibinfo{author}{L.~Ramos}, \bibinfo{author}{H.~G\'{o}mez-Adorno}, \bibinfo{author}{A.~Gelbukh},
\newblock \bibinfo{title}{{MIND-HOPE}: Multilingual identification of nuanced dimensions of {HOPE}}  (\bibinfo{year}{2024}).
\bibitem[{Balouchzahi et~al.(2025)Balouchzahi, Butt, Amjad, Sidorov, and Gelbukh}]{balouchzahi2025urduhope}
\bibinfo{author}{F.~Balouchzahi}, \bibinfo{author}{S.~Butt}, \bibinfo{author}{M.~Amjad}, \bibinfo{author}{G.~Sidorov}, \bibinfo{author}{A.~Gelbukh},
\newblock \bibinfo{title}{{UrduHope}: Analysis of hope and hopelessness in {Urdu} texts},
\newblock \bibinfo{journal}{Knowledge-Based Systems} \bibinfo{volume}{308} (\bibinfo{year}{2025}) \bibinfo{pages}{112746}.
\bibitem[{Gonz{\'a}lez-Barba et~al.(2025)Gonz{\'a}lez-Barba, Chiruzzo, and Jim{\'e}nez-Zafra}]{iberlef2025overview}
\bibinfo{author}{J.~{\'A}. Gonz{\'a}lez-Barba}, \bibinfo{author}{L.~Chiruzzo}, \bibinfo{author}{S.~M. Jim{\'e}nez-Zafra},
\newblock \bibinfo{title}{{Overview of IberLEF 2025: Natural Language Processing Challenges for Spanish and other Iberian Languages}},
\newblock in: \bibinfo{booktitle}{Proceedings of the Iberian Languages Evaluation Forum (IberLEF 2025), co-located with the 41st Conference of the Spanish Society for Natural Language Processing (SEPLN 2025), CEUR-WS. org}, \bibinfo{year}{2025}.
\bibitem[{Siino et~al.(2024)Siino, Tinnirello, and La~Cascia}]{siino2024text_preprocessing}
\bibinfo{author}{M.~Siino}, \bibinfo{author}{I.~Tinnirello}, \bibinfo{author}{M.~La~Cascia},
\newblock \bibinfo{title}{Is text preprocessing still worth the time? a comparative survey on the influence of popular preprocessing methods on transformers and traditional classifiers},
\newblock \bibinfo{journal}{Information Systems} \bibinfo{volume}{121} (\bibinfo{year}{2024}) \bibinfo{pages}{102342}. \URLprefix \url{https://doi.org/10.1016/j.is.2023.102342}. \DOIprefix\doi{10.1016/j. ::contentReference[oaicite:2]{index=2} is.2023.102342}.
\bibitem[{Turc et~al.(2020)Turc, Chang, Lee, and Toutanova}]{turc2019well}
\bibinfo{author}{I.~Turc}, \bibinfo{author}{M.-W. Chang}, \bibinfo{author}{K.~Lee}, \bibinfo{author}{K.~Toutanova},
\newblock \bibinfo{title}{Well-read students learn better: On the importance of pre-training compact models},
\newblock in: \bibinfo{booktitle}{Proceedings of the 8th International Conference on Learning Representations (ICLR)}, \bibinfo{publisher}{International Conference on Learning Representations}, \bibinfo{year}{2020}.
\bibitem[{Buechel and Hahn(2017)}]{buechel-hahn-2017-emobank}
\bibinfo{author}{S.~Buechel}, \bibinfo{author}{U.~Hahn},
\newblock \bibinfo{title}{{E}mo{B}ank: Studying the impact of annotation perspective and representation format on dimensional emotion analysis},
\newblock in: \bibinfo{editor}{M.~Lapata}, \bibinfo{editor}{P.~Blunsom}, \bibinfo{editor}{A.~Koller} (Eds.), \bibinfo{booktitle}{Proceedings of the 15th Conference of the {E}uropean Chapter of the Association for Computational Linguistics: Volume 2, Short Papers}, \bibinfo{publisher}{Association for Computational Linguistics}, \bibinfo{address}{Valencia, Spain}, \bibinfo{year}{2017}, pp. \bibinfo{pages}{578--585}. \URLprefix \url{https://aclanthology.org/E17-2092/}.
\bibitem[{Felbo et~al.(2017)Felbo, Mislove, S{\o}gaard, Rahwan, and Lehmann}]{felbo-etal-2017-using}
\bibinfo{author}{B.~Felbo}, \bibinfo{author}{A.~Mislove}, \bibinfo{author}{A.~S{\o}gaard}, \bibinfo{author}{I.~Rahwan}, \bibinfo{author}{S.~Lehmann},
\newblock \bibinfo{title}{Using millions of emoji occurrences to learn any-domain representations for detecting sentiment, emotion and sarcasm},
\newblock in: \bibinfo{editor}{M.~Palmer}, \bibinfo{editor}{R.~Hwa}, \bibinfo{editor}{S.~Riedel} (Eds.), \bibinfo{booktitle}{Proceedings of the 2017 Conference on Empirical Methods in Natural Language Processing}, \bibinfo{publisher}{Association for Computational Linguistics}, \bibinfo{address}{Copenhagen, Denmark}, \bibinfo{year}{2017}, pp. \bibinfo{pages}{1615--1625}. \URLprefix \url{https://aclanthology.org/D17-1169/}. \DOIprefix\doi{10.18653/v1/D17-1169}.
\bibitem[{Demszky et~al.(2020)Demszky, Movshovitz-Attias, Ko, Cowen, Nemade, and Ravi}]{demszky2020goemotions}
\bibinfo{author}{D.~Demszky}, \bibinfo{author}{D.~Movshovitz-Attias}, \bibinfo{author}{J.~Ko}, \bibinfo{author}{A.~Cowen}, \bibinfo{author}{G.~Nemade}, \bibinfo{author}{S.~Ravi},
\newblock \bibinfo{title}{Goemotions: A dataset of fine-grained emotions},
\newblock in: \bibinfo{booktitle}{Proceedings of the 58th Annual Meeting of the Association for Computational Linguistics}, \bibinfo{publisher}{Association for Computational Linguistics}, \bibinfo{year}{2020}, pp. \bibinfo{pages}{4040--4054}.
\bibitem[{Soleymani et~al.(2017)Soleymani, Garcia, Jou, Schuller, Chang, and Pantic}]{soleymani2017survey}
\bibinfo{author}{M.~Soleymani}, \bibinfo{author}{D.~Garcia}, \bibinfo{author}{B.~Jou}, \bibinfo{author}{B.~Schuller}, \bibinfo{author}{S.-F. Chang}, \bibinfo{author}{M.~Pantic},
\newblock \bibinfo{title}{A survey of multimodal sentiment analysis},
\newblock \bibinfo{journal}{Image and Vision Computing} \bibinfo{volume}{65} (\bibinfo{year}{2017}) \bibinfo{pages}{3--14}.
\bibitem[{Bollen et~al.(2011)Bollen, Mao, and Zeng}]{bollen2011twitter}
\bibinfo{author}{J.~Bollen}, \bibinfo{author}{H.~Mao}, \bibinfo{author}{X.~Zeng},
\newblock \bibinfo{title}{Twitter mood predicts the stock market},
\newblock \bibinfo{journal}{Journal of computational science} \bibinfo{volume}{2} (\bibinfo{year}{2011}) \bibinfo{pages}{1--8}.
\bibitem[{Nabi et~al.(2013)Nabi, Prestin, and So}]{nabi2013facebook}
\bibinfo{author}{R.~L. Nabi}, \bibinfo{author}{A.~Prestin}, \bibinfo{author}{J.~So},
\newblock \bibinfo{title}{Facebook friends with (health) benefits? exploring social network site use and perceptions of social support, stress, and well-being},
\newblock \bibinfo{journal}{Cyberpsychology, behavior, and social networking} \bibinfo{volume}{16} (\bibinfo{year}{2013}) \bibinfo{pages}{721--727}.
\bibitem[{MacInnis and De~Mello(2005)}]{macinnis2005concept}
\bibinfo{author}{D.~J. MacInnis}, \bibinfo{author}{G.~E. De~Mello},
\newblock \bibinfo{title}{The concept of hope and its relevance to product evaluation and choice},
\newblock \bibinfo{journal}{Journal of Marketing} \bibinfo{volume}{69} (\bibinfo{year}{2005}) \bibinfo{pages}{1--14}.
\bibitem[{Gururangan et~al.(2020)Gururangan, Marasovi{\'c}, Swayamdipta, Lo, Beltagy, Downey, and Smith}]{gururangan-etal-2020-dont}
\bibinfo{author}{S.~Gururangan}, \bibinfo{author}{A.~Marasovi{\'c}}, \bibinfo{author}{S.~Swayamdipta}, \bibinfo{author}{K.~Lo}, \bibinfo{author}{I.~Beltagy}, \bibinfo{author}{D.~Downey}, \bibinfo{author}{N.~A. Smith},
\newblock \bibinfo{title}{Don{'}t stop pretraining: Adapt language models to domains and tasks},
\newblock in: \bibinfo{editor}{D.~Jurafsky}, \bibinfo{editor}{J.~Chai}, \bibinfo{editor}{N.~Schluter}, \bibinfo{editor}{J.~Tetreault} (Eds.), \bibinfo{booktitle}{Proceedings of the 58th Annual Meeting of the Association for Computational Linguistics}, \bibinfo{publisher}{Association for Computational Linguistics}, \bibinfo{address}{Online}, \bibinfo{year}{2020}, pp. \bibinfo{pages}{8342--8360}. \URLprefix \url{https://aclanthology.org/2020.acl-main.740/}. \DOIprefix\doi{10.18653/v1/2020.acl-main.740}.
\bibitem[{Jackson et~al.(2019)Jackson, Watts, Henry, List, Forkel, Mucha, Greenhill, Gray, and Lindquist}]{jackson2019emotion}
\bibinfo{author}{J.~C. Jackson}, \bibinfo{author}{J.~Watts}, \bibinfo{author}{T.~R. Henry}, \bibinfo{author}{J.-M. List}, \bibinfo{author}{R.~Forkel}, \bibinfo{author}{P.~J. Mucha}, \bibinfo{author}{S.~J. Greenhill}, \bibinfo{author}{R.~D. Gray}, \bibinfo{author}{K.~A. Lindquist},
\newblock \bibinfo{title}{Emotion semantics show both cultural variation and universal structure},
\newblock \bibinfo{journal}{Science} \bibinfo{volume}{366} (\bibinfo{year}{2019}) \bibinfo{pages}{1517--1522}.
\bibitem[{Sun et~al.(2019)Sun, Qiu, Xu, and Huang}]{sun2019fine}
\bibinfo{author}{C.~Sun}, \bibinfo{author}{X.~Qiu}, \bibinfo{author}{Y.~Xu}, \bibinfo{author}{X.~Huang},
\newblock \bibinfo{title}{How to fine-tune bert for text classification?},
\newblock in: \bibinfo{booktitle}{China national conference on Chinese computational linguistics}, \bibinfo{organization}{Springer}, \bibinfo{year}{2019}, pp. \bibinfo{pages}{194--206}.
\bibitem[{Zhang et~al.(2021)Zhang, Bengio, Hardt, Recht, and Vinyals}]{ZhangUnderstanding}
\bibinfo{author}{C.~Zhang}, \bibinfo{author}{S.~Bengio}, \bibinfo{author}{M.~Hardt}, \bibinfo{author}{B.~Recht}, \bibinfo{author}{O.~Vinyals},
\newblock \bibinfo{title}{Understanding deep learning (still) requires rethinking generalization},
\newblock \bibinfo{journal}{Commun. ACM} \bibinfo{volume}{64} (\bibinfo{year}{2021}) \bibinfo{pages}{107–115}. \URLprefix \url{https://doi.org/10.1145/3446776}. \DOIprefix\doi{10.1145/3446776}.
\bibitem[{Richards and King(2014)}]{richards2014big}
\bibinfo{author}{N.~M. Richards}, \bibinfo{author}{J.~H. King},
\newblock \bibinfo{title}{Big data ethics},
\newblock \bibinfo{journal}{Wake Forest L. Rev.} \bibinfo{volume}{49} (\bibinfo{year}{2014}) \bibinfo{pages}{393}.
\bibitem[{Crawford and Calo(2016)}]{crawford2016there}
\bibinfo{author}{K.~Crawford}, \bibinfo{author}{R.~Calo},
\newblock \bibinfo{title}{There is a blind spot in ai research},
\newblock \bibinfo{journal}{Nature} \bibinfo{volume}{538} (\bibinfo{year}{2016}) \bibinfo{pages}{311--313}.
\bibitem[{Floridi et~al.(2018)Floridi, Cowls, Beltrametti, Chatila, Chazerand, Dignum, Luetge, Madelin, Pagallo, Rossi et~al.}]{floridi2018ai4people}
\bibinfo{author}{L.~Floridi}, \bibinfo{author}{J.~Cowls}, \bibinfo{author}{M.~Beltrametti}, \bibinfo{author}{R.~Chatila}, \bibinfo{author}{P.~Chazerand}, \bibinfo{author}{V.~Dignum}, \bibinfo{author}{C.~Luetge}, \bibinfo{author}{R.~Madelin}, \bibinfo{author}{U.~Pagallo}, \bibinfo{author}{F.~Rossi}, et~al.,
\newblock \bibinfo{title}{Ai4people—an ethical framework for a good ai society: opportunities, risks, principles, and recommendations},
\newblock \bibinfo{journal}{Minds and machines} \bibinfo{volume}{28} (\bibinfo{year}{2018}) \bibinfo{pages}{689--707}.

\end{thebibliography}

\section{Terminology}

This appendix provides definitions of specialized terms used throughout the paper that may not be familiar to all readers.

\begin{description}
    \item[BERT] Bidirectional Encoder Representations from Transformers. A transformer-based machine learning model for natural language processing pre-trained on a large corpus of text.

    \item[GPT-2] Generative Pre-trained Transformer 2. An autoregressive language model that uses unidirectional attention (each token can only attend to previous tokens). It contains 124 million parameters in its base version and was pre-trained on a larger corpus than BERT, but its unidirectional nature may limit contextual understanding for classification tasks.

    \item[DeBERTa] Decoding-enhanced BERT with Disentangled Attention. A transformer model that implements a novel attention mechanism which separately computes attention weights for content and position information. This architecture aims to provide more nuanced contextual understanding by disentangling the content and position information in the self-attention mechanism.
    
    \item[bert-base-uncased] A specific pre-trained variant of BERT that uses a vocabulary of uncased (lowercase) text. It contains 12 transformer layers, 12 attention heads, and 110 million parameters.
    
    \item[Generalized Hope] A broad, non-specific form of hope that is not tied to a particular outcome, timeframe, or realistic expectation. Often expressed as general optimism about the future.
    
    \item[Realistic Hope] Hope that is grounded in reality, with reasonable expectations of what could potentially happen based on evidence, experience, or logical reasoning.
    
    \item[Unrealistic Hope] Hope characterized by expectations that have a very low probability of being realized, often disregarding evidence or practical limitations.
    
    \item[Sarcasm] In the context of hope classification, expressions that superficially appear hopeful but actually convey the opposite meaning through irony, often with the intent to mock or criticize.
    
    \item[Fine-tuning] The process of taking a pre-trained model (like BERT) and further training it on a specific task or domain with a smaller dataset to adapt its knowledge to that particular application.
    
    \item[Attention Masks] Binary tensors used in transformer models to indicate which tokens should be attended to and which should be ignored (such as padding tokens).
    
    \item[Transfer Learning] A machine learning technique where knowledge gained while solving one problem is applied to a different but related problem, often allowing models to perform well with less task-specific data.
    
    \item[Tokenization] The process of breaking text into smaller units called tokens, which could be words, subwords, or characters, that serve as the input to NLP models.
    
    \item[Transformer Architecture] A deep learning architecture that uses self-attention mechanisms to process sequential data, allowing the model to weigh the importance of different words in relation to each other regardless of their position in the sequence.
    
    \item[TFBertForSequenceClassification] A TensorFlow implementation of BERT specifically designed for sequence classification tasks, with an additional classification layer on top of the BERT model.
    
    \item[SparseCategoricalCrossentropy] A loss function used in multi-class classification problems when the target values are represented as integers rather than one-hot encoded vectors.
    
    \item[Legacy Adam Optimizer] A version of the Adam optimization algorithm in TensorFlow that maintains compatibility with older implementations. Adam (Adaptive Moment Estimation) combines the benefits of two other extensions of stochastic gradient descent: AdaGrad and RMSProp.
    
    \item[Learning Rate] A hyperparameter that controls how much to change the model in response to the estimated error each time the model weights are updated. The value 2e-5 (0.00002) is commonly used for fine-tuning BERT models.
    
    \item[ModelCheckpoint Callbacks] Functions in TensorFlow that save the model's state at specific points during training, typically when the model achieves better performance on validation data than it has previously.
    
    \item[TensorFlow Format] A file format for saving TensorFlow models that preserves the model architecture, weights, and computational graph, allowing for model reuse and deployment.
    
    \item[Weighted Metrics] Performance metrics (precision, recall, F1-score) that account for class imbalance by calculating scores for each class and then taking a weighted average based on the number of samples in each class.
    
    \item[Macro Metrics] Performance metrics that calculate scores for each class independently and then take an unweighted average, treating all classes equally regardless of their size.
    
    \item[F1-Score] A measure of a model's accuracy that combines precision and recall. It is the harmonic mean of precision and recall, providing a balance between the two metrics.
    
    \item[Overfitting] A modeling error that occurs when a model learns the training data too well, including its noise and outliers, resulting in poor performance on new, unseen data.
    
    \item[Epoch] One complete pass through the entire training dataset during the training of a machine learning model.
\end{description}

\end{document}